\documentclass[10pt,twocolumn,letterpaper]{article}

 \usepackage{cvpr}              %

\definecolor{cvprblue}{rgb}{0.21,0.49,0.74}
\usepackage[pagebackref,breaklinks,colorlinks,allcolors=cvprblue]{hyperref}

\def\paperID{30} %
\def\confName{CVPR}
\def\confYear{2025}

\title{Investigating Different Geo Priors for Image Classification}

\author{Angela Zhu\textsuperscript{1}, Christian Lange\textsuperscript{2}, Max Hamilton\textsuperscript{1}\\
\\
\textsuperscript{1}University of Massachusetts Amherst\\
\textsuperscript{2}The University of Edinburgh\\
}

\begin{document}
\maketitle
\begin{abstract}
Species distribution models encode spatial patterns of species occurrence making them effective priors for vision-based species classification when location information is available. In this study, we evaluate various SINR (Spatial Implicit Neural Representations) models as a geographical prior for visual classification of species from iNaturalist observations. We explore the impact of different model configurations and adjust how we handle predictions for species not included in Geo Prior training. Our analysis reveals factors that contribute to the effectiveness of these models as Geo Priors, factors that may differ from making accurate range maps.
\end{abstract}
    
\section{Introduction}
\label{sec:intro}
Ecological research can generate vast amounts of data through methods such as camera traps, or by leveraging community collected data.
However, analyzing this data and manually labeling species is time-consuming and difficult to scale \cite{beery2021species}. 
As a result, integrating computational methods to automate and enhance species classification from images is a useful tool for ecological studies. 
In order to assist in research, vision models must be capable of fine-grained categorizations, identifying specific species rather than coarser classifications at the genus or family level. 

While vision models rely solely on visual features from an image to classify species, biologists tend to rely on additional contextual information such as location and time. 
Geographical information, in particular, can help disambiguate between visually similar species that occupy different regions.
By combining predictions from vision models with geospatial models, we can leverage location data to improve classification accuracy beyond what traditional vision-only approaches can achieve.

This method of combining the predictions of the two models also enables independent training and updating of image classification models and geospatial models. Since training vision models is computationally intensive and time-consuming, this separation allows for flexible updates to either component without retraining the entire system. This has been found to be effective on the iNaturalist platform which uses a ``geomodel" \cite{inaturalist_a, inaturalist_b} to assist in visual classification of species observations and significantly enhances classification performance.
Another advantage of this approach is that it eliminates the need for paired image-location datasets, allowing the models to be trained separately on images without location information and on location data without corresponding images.

The motivation behind this study is to explore the extent to which geospatial models can assist in fine-grained visual classification of species.
Specifically we are focusing on a deep learning based species distribution model that was introduced in Cole et al's work Spatial Implicit Neural Representation (SINR) \cite{cole2023spatial} which has been explored as a method of improving visual classification tasks.

In this study we investigate factors that can improve the performance of a deep species distribution model ``SINR" (Spatial Implicit Neural Representation) for use as a Geo Prior. This model is similar to the geomodel used by iNaturalist. 
We examine the impact of the amount of training, size of the model, modifications to the output probabilities and a method of handling species not seen during Geo Prior training. We provide a detailed evaluation of these modifications of the SINR model using the benchmark suite of evaluations introduced in \cite{cole2023spatial}.
By evaluating these factors, we aim to advance the integration of geospatial data into visual classification of wildlife images.

 \section{Related Works}
Incorporating geographical information into vision-based classification has been explored in several recent works. A notable example of geospatial model integration in visual species classification was introduced by \citet{mac2019presence} using a deep neural network as a spatio-temporal prior that estimates the probability of an object category occurring at a given location and time. Trained on presence-only observation data, their model jointly captures object categories, their spatio-temporal distributions, and photographer biases. By combining this prior with image classifier predictions, they achieved significant improvements in classification performance. 

\citet{dorm2024generating} examined the impact of binarizing SDM outputs into presence-absence maps, finding that thresholded maps generally performed worse than unthresholded probabilistic distributions. They also introduced a variant that assigns small baseline probabilities to all locations ensuring no locations have zero probability for any species, yielding minor improvements. 
Incorporating geographical information into vision-based classification has been explored in several recent works.
\cite{yang2022dynamic} introduced the Dynamic MLP, their approach demonstrated that incorporating spatial priors significantly improves classification accuracy over standard vision models and non-neural alternatives such as nearest neighbor and adaptive kernel methods.
A complementary study, Sat-SINR \cite{isprs-annals-X-2-2024-41-2024}, explores how satellite imagery can enhance species distribution models through late fusion, where separate models process location and image data independently before their predictions are combined. This setup aligns with the "Geo Prior" task design, where location information is used as an auxiliary input rather than being fused early into the vision model. The late fusion approach is advantageous because it allows for independent updates to either the image encoder or location encoder, reduces training costs, and ensures that either modality can function independently when only one source of data is available. In contrast, early fusion methods often struggle when partial input data is missing.

These studies collectively highlight the effectiveness of integrating spatial priors into vision tasks and provide motivation for flexible, modular approaches to incorporating geographical information in classification pipelines. 
Many approaches which integrate geospatial and visual information integrate these into a single model, which can be expensive to train and requires paired location - image datasets.

\section{Method}
\label{sec:experiments}

\begin{table*}[t]
    \centering
    \resizebox{0.55\textwidth}{!}{%
    \begin{tabular}{lll|lll}
        & Method & Parameters & IUCN & S\&T & GeoPrior\\ \hline
        1 & Vision Only &  & -- & -- & 75.409\\
        2 & Baseline SINR & 10 epochs, k =1.0 & 62.00 & 72.82 &  +6.6\\ \hline
        3 & Baseline SINR & 20 epochs, delta = 0.001 & 64.35 & 73.58 & +6.77\\
        4 & Baseline SINR & 10 epochs, k = 0.02 & 62.00 & 72.78 & \textbf{+9.0}\\
    \end{tabular}
    \vspace{-10pt}
    }
    \caption{\textbf{Overall results for geospatial tasks}. IUCN (IUCN species mapping), S\&T (eBird Status and Trends species mapping), Geo Prior (fine-grained image classification with a geographical prior). The baseline SINR model (1) and vision-only model (2) are from \cite{cole2023spatial}. The baseline/default SINR configuration uses an full with observations capped at 100 and is trained for 10 epochs. Model 3 is built off of baseline SINR, it is trained for 10 more epochs and has a delta of 0.01, method described in \ref{sec:def_delta}. Model 4 is also build off of baseline SINR but incorporates a k parameter which adjusts how we handle non-SINR species defined at \ref{sec:def_kparam}.}
    \label{tab:eval_table}
\end{table*}
Through this investigation, we aim to refine the application of geomodels in fine-grained visual classification and contribute insights into the distinct modeling requirements for Geo Priors versus range mapping tasks. Our focus will be on Spatial Implicit Neural Representations (SINRs) \cite{cole2023spatial}, which is a multi-species neural network-based species distribution model that is trained on presence-only species observation data collected from the citizen science platform iNaturalist. This model is tasked with predicting species presence based on a location and can be used to generate global species distributions. 

\noindent The two components of our model are
\begin{enumerate}
    \item \textbf{Geographic Model}: Predicts species presence probability based solely on location, estimating $P(species|location)$
    \item \textbf{Image Model}: Produces species probability estimates from image inputs, computing $P(species|image)$.
\end{enumerate}

At test time, the input into our SINR-based model is the location of an observation from iNaturalist represented as a longitude and latitude coordinate, and the output is a vector representing the predicted probabilities of presence at that location for each of the 42,225 species our model is trained on. The input into the vision model is the image associated with each observation and the output is a vector of predictions of species likelihood for all species the vision model can predict on. The final classification probability is obtained by multiplying these independent probabilities for each species.
\begin{equation}
    \resizebox{\columnwidth}{!}{%
    $P(species|(image, location)) \approx P(species|location)*P(species|image)$
    }
\end{equation}
While image features may indirectly contain location-related cues (e.g., background scenery), prior studies indicate that assuming independence between the two models does not significantly degrade performance \cite{inaturalist_a}.

We hypothesize that modifying the SINR species distribution model to produce larger, smoother range estimates will improve GeoPrior classification performance. This change is motivated by the limitations of citizen science data, which may not fully capture a species' range. By slightly overestimating species presence in ambiguous regions, we aim to enhance recall and make the Geo Prior more effective, even when species are observed outside their traditional habitats. We tested three methods: adjusting training epochs, changing the dimensions of hidden layers, and adding a delta to all species. These changes are expected to reduce overfitting and result in smoother predictions, which could improve GeoPrior performance, especially with lower dimensions and fewer training epochs.
\subsection{Evaluation Tasks}
\label{deftasks}
Our primary evaluation focuses on measuring the utility of SDMs as geographical priors (Geo Priors) for fine-grained image classification, which is built off of work from \citet{mac2019presence}. 
Additionally we assess how effective our range maps measure against expert-derived maps per species.
Unless otherwise specified, all baseline models in this study are based on the best-performing configuration from \citet{cole2023spatial}. This setup employs the full assume negative loss, a pseudo-sampling method for negative examples. Training is limited to a maximum of 100 observations per class, a neural network with 256 hidden dimensions and running for 10 epochs.

\noindent \textbf{Image Classification: Geo Prior} We use the same evaluation data set from \cite{mac2019presence, cole2023spatial} in the Geo Prior task, that is comprised of 282,974 images of species from iNaturalist.  The evaluation metric is the Top-1 image classification accuracy resulting from combining range predictions with the probabilistic classification outputs of an Xception classifier trained on images from iNaturalist, following the methodology described in \citet{cole2023spatial}. This task is measured as a percentage of gain that combined vison model and geomodel get over classification performance of just the vision model itself. 

\noindent \textbf{Range Map Accuracy: S\&T and IUCN.} We measure performance of the geomodel's range predictions against expert-derived range maps from the eBird Status \& Trends (S\&T) dataset (\citet{fink2020ebird}) and the International Union for Conservation of Nature (IUCN) Red List. Performance is measured using mean average precision (MAP) which computes the per-species average precision then averages over all species. Following \citet{cole2023spatial}, we use the S\&T dataset with 535 bird species and the IUCN Expert Range Maps dataset with 2,418 species across different taxonomic groups.

\section{Results}
\label{sec:results}

\subsection{Attempt at Larger and Smoother Range Estimations}
The next three sections outline our approach to adjusting SINR model configurations to better align with our desired predictive behavior. We strive to make modifications that enhance Geo Prior performance by reducing restrictive constraints on location-based predictions. However, this approach could also lead to a tradeoff in performance for stricter evaluation benchmarks, such as S\&T and IUCN metrics, where range accuracy is prioritized over flexibility. 

\noindent \textbf{Effect of Training Duration}
We experimented with training the model for different levels of epochs from 5 to 50, saving models at every 5 epochs. Training for fewer epochs than default led to worse performance in the Geo Prior task. Conversely, longer training durations generally improved performance, but there was a sweet spot at approximately 20 epochs for default SINR, beyond which gains plateaued or diminished slightly. The overall gain is minimal at 0.1\% and could easily be explained by random noise.

\begin{table}[t]
    \centering
    \resizebox{.75\columnwidth}{!}{%
    \begin{tabular}{ll|llll}
        Method & Epochs & S\&T & IUCN & GeoPrior & +Delta\\ \hline
        Vision Only & & -- & -- & 75.409 & 75.409\\  
        SINR & 10 & 62.00 & 72.82 & +6.6 & +6.58\\ \hline
        SINR & 5 & 59.42 & 72.28 & +6.26 & +6.29\\
        SINR & 20 & 64.35 & 73.58 & \textbf{+6.68} & \textbf{+6.77}\\
        SINR & 30 & 64.75 & \textbf{73.60} & +6.61 & +6.75\\
        SINR & 40 & 65.65 & 73.42 & +6.59 & +6.75\\
        SINR & 50 & \textbf{65.77} & 73.36 & +6.57 & +6.75\\
    \end{tabular}
    }
    \caption{\textbf{Varying Training Epochs and adding Delta.} All models are based on the baseline SINR, with variations in the number of training epochs and the addition of a delta adjustment. The bold values are the best performing for each task. Adding a delta of 0.001 consistently provided a slight performance boost across all cases. Increasing the number of training epochs generally improved performance, with the best GeoPrior performance observed at 20 epochs. Performance values for the GeoPrior task are reported as the improvement over the vision-only model.}
    \label{tab:trainingtable}
\end{table}

\noindent \textbf{Smaller vs Larger Models}
We experimented with modifying the number of dimensions of the hidden layers of the neural network in SINR in an effort to control the complexity of representations the model is producing. Reducing the size of hidden layers effectively coarsened the model, making its predictions more spatially generalized and coarse rather than highly detailed. The baseline SINR model used 256 filters, and we tested configurations with [64, 128, 256, 512, 1024] filters to assess their impact on model performance. 

Our results in Table \ref{tab:smallvlarge} indicate that smaller SINR models required more training epochs to achieve comparable performance to larger models. In contrast, larger SINR models converged to an effective Geo Prior faster, requiring fewer epochs to reach similar results. Both slightly outperforming the default SINR configuration by approximately 0.1\%, at such a small improvement this could again be explained by random noise. Ultimately, varying model size does not outperform default SINR that is trained for more epochs.
\begin{table}[t]
    \centering
    \resizebox{\columnwidth}{!}{%
    \begin{tabular}{lll|llll}
        Method & Hidden Dim & \#Epochs & S\&T & IUCN & GeoPrior & +Delta\\ \hline
        Vision Only & & & -- & -- & 75.409 & 75.409\\  
        Default SINR & 256 & 10 & 62.00 & 72.82 & +6.6 & +6.58\\ \hline
        Even Smaller SINR & 64 & 50 & 61.24 & 72.28 & +6.32 & +6.41\\
        Smaller SINR & 128 & 50 & 64.35 & \textbf{74.03} & +6.65 & +6.73\\
        Baseline SINR & 256 & 20 & 64.35 & 73.58 & \textbf{+6.68} & \textbf{+6.77}\\
        Bigger SINR & 512 & 10 & 63.95 & 73.64 & +6.64 & +6.64\\
        Even Bigger SINR & 1024 & 15 & \textbf{66.85} & 72.48 & +6.49 & +6.71 
    \end{tabular}
    }
    \caption{\textbf{Effect of Model Size and adding Delta.} This table presents the best-performing models across different hidden dimension sizes and training epochs. All models use a baseline SINR configuration, with variations in model size and training duration.The Delta column reflects performance when a delta of 0.001 was added to the Geo Prior task, with performance reported as the improvement over the vision-only model. Bold values indicate the best-performing results for each task. As shown, varying model size has minimal impact on overall performance.}
    \label{tab:smallvlarge}
\end{table}
\noindent \textbf{Adding Delta to All Species Predictions}
\label{sec:def_delta}
This approach was explored in \citet{dorm2024generating}, where incorporating a delta value to all species, thereby ensuring a minimum prediction at every location, improved performance over using only the raw species distribution model predictions. 
The best-performing delta value across different model configurations was 0.001, results for which are shown in Tables \ref{tab:trainingtable}, \ref{tab:smallvlarge}, followed closely by 0.005. 
Both outperformed using no delta adjustment or a higher delta value of 0.1, with an overall improvement of 0.1\% compared to default SINR. These results confirm the findings in the original paper \cite{dorm2024generating}.

\subsection{Handling Unseen Species}
\label{sec:def_kparam}
\begin{table}[t]
    \centering
    \resizebox{0.60\columnwidth}{!}{%
    \begin{tabular}{ll|lll}
        Method & K Value & GeoPrior\\ \hline
        Vision Only & & 75.409\\  
        Baseline SINR & 1.0 & +6.6\\ \hline
        Baseline SINR & 0.01 & +8.9\\
        Baseline SINR & 0.02 & \textbf{+9.0}\\
        Baseline SINR & 0.1 & +8.7\\
        Baseline SINR & 0.2 & +8.3\\
    \end{tabular}
    }
    \caption{\textbf{Effect of K-Value on Handling Unseen Species.} This table shows the impact of varying the K-value, which controls the default probability assigned to species that the SINR model cannot predict but the vision model can classify. All models are based on the baseline SINR configuration. Bold values indicate the best-performing result. The baseline SINR and vision-only models are from \citet{cole2023spatial}. Adjusting K significantly improved performance, with the best result (+9.0) at K = 0.02, yielding a 2.4\% improvement over default SINR—the highest gain observed across our experiments.}
    \label{tab:k}
\end{table}

There is a gap between the species that SINR can predict and the species that only the vision model can classify. Specifically, SINR is trained to make predictions for 42,225 species, while the vision model can recognize 55,378 species, leaving 13,153 species that SINR cannot predict. 
To address this, we introduce a parameter K, which defines the default probability assigned to species not predicted by SINR in the location model's predictions. Initially, K was set to 1.0 to prevent the SINR model from influencing the vision model’s predictions for species it had never encountered during training.
However, this caused an imbalance. Since all SINR-predicted probabilities were inherently less than 1.0, this down-weighted SINR species compared to non-SINR species. Even for species present in the training set and observed close to a test location, their probability for that location would likely be less than 1.0, making species with no geographic data appear more likely to be present than species observed nearby. To address this, we performed a grid search to determine the optimal value for K.
As shown in Table \ref{tab:k}, this adjustment resulted in the highest improvement in the GeoPrior category, with a 2.4\% gain over the baseline SINR model. In comparison, our other methods only contributed a 0.1\% improvement. This result highlights the importance of carefully handling unseen species when combining two models to ensure expected behavior.

\section{Conclusion}
\label{sec:conclusion}
In this study, we explored methods to enhance species distribution models as geographical priors (Geo Priors) for fine-grained visual classification. Our goal was to improve the accuracy and reliability of species range estimations used in vision-based models. We experimented with adjustments to SINR training parameters, including the number of training epochs, hidden dimensions, and a delta adjustment, to assess their impact on Geo Prior performance. Additionally, we examined how handling species not covered by the SINR model affected overall classification results.

Training SINR for fewer epochs resulted in worse Geo Prior performance, while longer training improved results, with the optimal balance occurring around 20 epochs. The delta adjustment of 0.001 consistently provided a small improvement in Geo Prior performance. Adjusting the number of hidden dimensions in the neural network did not improve performance. Training the models for longer led to an increase in performance with our best-performing model was the default SINR trained for 20 epochs instead of the original 10, along with the addition of a delta of 0.001. However, these changes resulted in only moderate performance improvements, with a total increase of 0.1\%.

The most significant improvement came from refining how we handled combining the two models and assigning a value to the species that the species distribution model could not predict but the vision model could.
Modifying how non-SINR species are handled yielded had the largest impact, improving performance in the Geo Prior category by up to 2.4\% over baseline SINR. We updated the location prediction from SINR to be 0.02.

For future directions, instead of setting unseen species to a static value we could use a dynamic value, for example the average prediction across all species at a given location. We could run the same evaluation on other datasets to see what k-value performed best. We could try zero-shot techniques in addition to SINR to improve performance on species that SINR has not seen before. A great candidate for this would be Language Enhanced SINR (LE-SINR) by \citet{hamilton2024}. This method performs worse compared to SINR for species SINR by \citet{cole2023spatial} has been trained to predict for but it has the capability to make zero shot predictions. LE-SINR is able to predict for any species if it is given text on that species. 

{
    \small
    \bibliographystyle{ieeenat_fullname}
    \bibliography{main}
}

\end{document}